\documentclass[letterpaper, 10 pt, conference]{ieeeconf}  

\usepackage{amsmath}
\usepackage{graphicx}
\usepackage{color}
\usepackage[table]{xcolor}
\usepackage{amsmath,amsfonts}
\usepackage{algorithmic}
\usepackage{algorithm}
\usepackage{array}
\usepackage[caption=false,font=normalsize,labelfont=sf,textfont=sf]{subfig}
\usepackage{textcomp}
\usepackage{stfloats}
\usepackage{url}
\usepackage{verbatim}
\usepackage{graphicx}
\usepackage{cite}
\usepackage{bm}
\usepackage{amsfonts}
\usepackage{multirow}
\usepackage{multicol}
\usepackage{booktabs}
\usepackage{algorithmic}
\usepackage{algorithm}
\usepackage{makecell}
\usepackage[colorlinks,linkcolor=blue]{hyperref}
\DeclareUnicodeCharacter{00A0}{~}
\IEEEoverridecommandlockouts                              

\title{\LARGE \bf
OVGNet: A Unified Visual-Linguistic Framework for Open-Vocabulary Robotic Grasping }

\author{Meng Li$^{1, 2}$, Qi Zhao$^{1}$, Shuchang Lyu$^{1}$, Chunlei Wang$^{1}$, Yujing Ma$^{3}$, Guangliang Cheng$^{4}$ and Chenguang Yang$^{4}$
\thanks{Corresponding author: Guangliang Cheng} 
\thanks{*This work was supported by the National Natural Science Foundation of China (grant number 62072021)}
\thanks{$^{1}$Meng Li, Qi Zhao, Shuchang Lyu and Chunlei Wang are with Department of Electronic and Information Engineering, Beihang University. \{limenglm, zhaoqi, lyushuchang, wcl\_buaa\}@buaa.edu.cn}%
\thanks{$^{2}$Meng Li is also with Peng Cheng Laboratory, ShenZhen, China}%
\thanks{$^{3}$Yujing Ma is with SenseTime, Beijing, China. mayujing@senseauto.com}%
\thanks{$^{4}$Guangliang Cheng and Chenguang Yang are with the Department of Computer Science, at the University of Liverpool. Guangliang.Cheng@liverpool.ac.uk, cyang@ieee.org}%
}

\begin{document}

\maketitle
\thispagestyle{empty}
\pagestyle{empty}

\begin{abstract}
\par Recognizing and grasping novel-category objects remains a crucial yet challenging problem in real-world robotic applications. Despite its significance, limited research has been conducted in this specific domain. To address this, we seamlessly propose a novel framework that integrates open-vocabulary learning into the domain of robotic grasping, empowering robots with the capability to adeptly handle novel objects. Our contributions are threefold. Firstly, we present a large-scale benchmark dataset specifically tailored for evaluating the performance of open-vocabulary grasping tasks. Secondly, we propose a unified visual-linguistic framework that serves as a guide for robots in successfully grasping both base and novel objects. Thirdly, we introduce two alignment modules designed to enhance visual-linguistic perception in the robotic grasping process. Extensive experiments validate the efficacy and utility of our approach. Notably, our framework achieves an average accuracy of 71.2\% and 64.4\% on base and novel categories in our new dataset, respectively. Our code and dataset are available at \href{https://github.com/cv516Buaa/OVGNet}{https://github.com/cv516Buaa/OVGNet}.

\end{abstract}
\section{Introduction}
\label{sec:intro}
\par The increasing demand for intelligent robots in diverse real-world applications, including warehousing logistics~\cite{Warehousing-grasp}, domestic service~\cite{Domestic-grasp}, and smart agriculture~\cite{Agriculture-grasp}, exposes limitations in current robotic perception and grasping methods. Existing methods are limited to recognizing predefined (seen) categories, hindering their ability to handle novel-category (unseen) objects in real-world scenarios where listing all possible categories is impractical. Hence, it is highly necessary to introduce an algorithm for detecting and grasping novel-category objects in the field of robotics.
\par Recent advancements in the field of visual-linguistic robotic grasping have facilitated human-guided interaction with robots through language. Various methods~\cite{VL-Grasp,VLAGrasp,Tang_GraspGPT} have been proposed to develop deep learning-based frameworks that address the challenge of visual-linguistic feature representation. These frameworks empower robots to execute object grasping tasks based on language guidance. However, there still exists two main weaknesses in visual-linguistic robotic grasping task. First, previous methods overlook the representation potential on localizing and grasping novel (unseen) objects. The benchmark datasets on this task is also lacking. Second, with the development of open-vocabulary learning (OVL), some notable methods~\cite{CLIP, Grounding-DINO} provide an insight on detecting novel objects. However, a unified framework, which integrates OVL mechanism into robotic grasping still remains unexplored. 
\begin{figure}
  \centering
  \includegraphics[width=1.0\linewidth]{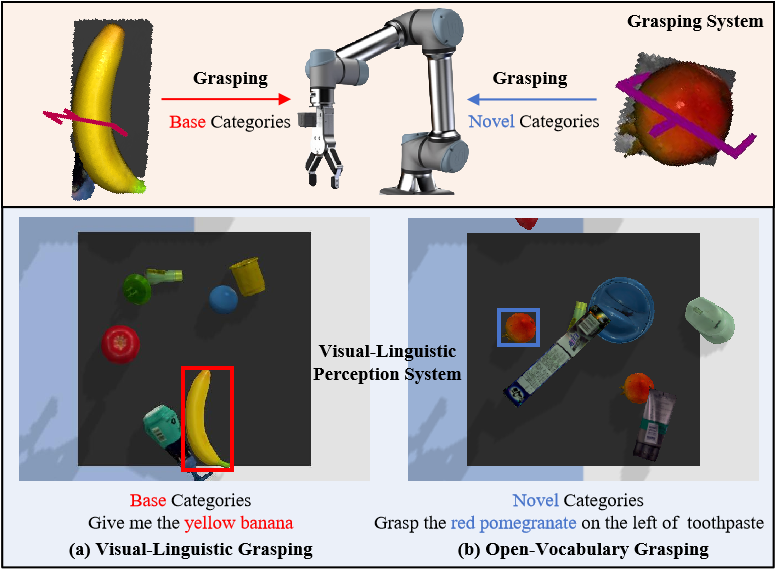}     \\
  \caption{\textbf{The diagram of open-vocabulary grasping.} Objects and fonts in red and blue respectively indicate the base and novel categories.}
  \label{Fig1}
\end{figure}
\par To address the aforementioned weaknesses, we introduce an innovative benchmark dataset named OVGrasping specifically tailored for the open-vocabulary grasping (OVG) task. Comprising 63,385 instances across 117 categories, the dataset is strategically partitioned into base and novel categories. The primary objective is to guide robotic systems in recognizing and effectively grasping novel objects. Additionally, we also introduce the Open-Vocabulary Learning (OVL) concept into the realm of robotic grasping, presenting a comprehensive framework termed as OVGNet for the OVG task. The framework focus on localizing and grasping novel objects by leveraging the learned knowledge from open-vocabulary foundation models and fine-tuning knowledge with only base categories. As shown in Fig.~\ref{Fig1}, our framework comprises a visual-linguistic perception system for locating target objects based on language references and a grasping system for acquiring them using specific grasping poses. To improve the perceptual capability for novel objects, we introduce an image guided language attention module (IGLA) and a language guided image attention module (LGIA) within the visual-linguistic system. These modules ensure alignment between visual and linguistic features, thereby promoting generalization from base-category to novel-category objects. Extensive experiments conducted on the OVGrasping dataset validate the efficacy of OVGNet. Visualization and analysis demonstrate the interpretability of our method.
\par Comparing to previous methods (Fig.~\ref{Fig1}). The VLG task focuses on locating and grasping base-category objects, which may fail when encountering with novel-category objects. In contrast, the OVG task aims to generalize the locating and grasping ability, consequently facilitating open-vocabulary grasping for novel-category objects. In summary, the main contributions are listed as follows: 
\begin{itemize}
\item We are the first to introduce a large-scale dataset for benchmarking the OVG task, providing a valuable resource for guiding robots in grasping both base and novel categories.
\item We propose a novel OVG framework that integrates the OVL into robotic grasping, which leverages both prior knowledge and fine-tuning techniques, significantly enhancing the robot's ability to grasp novel-category objects. 
\item We introduce two alignment modules to enhance feature consistency, thereby improving the generalization capabilities of our visual-linguistic perception system.
\end{itemize}
\section{Related Works}
\label{sec:rw}
\subsection{Grasp Pose Detection}
\par Grasping pose detection mainly falls into two categories. 2D grasping~\cite{2Dgrasp} detects the grasping points and orientations within a two-dimensional plane. In contrast, 6-DOF grasping~\cite{RGBD-Grasp,Monograspnet,Graspnerf,Cheng-TIM-RL} is the most extensively method. GraspNet~\cite{GraspNet} has constructed a dataset containing one billion grasp poses and provides a baseline for detecting grasp poses. Furthermore, Lu et al.~\cite{FGC-GraspNet} propose the FGC-Grasp which integrates force-closure with measurements of flatness, gravity, and collision. Xu et al.~\cite{Xu-ICRA-RL} propose a single-stage grasping network to execute instance-level grasping. Wang et al.~\cite{Wang-RAL-RL} propose a visual grasping framework based transformer, which captures both local and global features. 

\subsection{Open-Vocabulary Visual Grounding}

\par Open-Vocabulary learning~\cite{Li-ECCV-2020,Li-TIP-2021} aims to expand the range of vocabulary and comprehension. Previous works mainly involved two aspects: Open-Vocabulary Detection (OV-D)\cite{Rasheed-NIPS-2022,OVD-Du-CVPR-2022,Zhang-CVPR-OVSD,Wu-TCSVT-2022} and Open-Vocabulary Segmentation (OV-S)~\cite{OVS-Ma-BMVC-2022,OVS-Liang-CVPR-2023,Wu-ICCV-2023,Xu-CVPR-2023}. OV-D methods aim to detect the objects without the predefined category. These approaches can accurately detect various objects instructed by language. Liu et al.~\cite{Grounding-DINO} propose an OV-D framework based on  DINO~\cite{Dino}, which outputs the score between phrases and objects. Unlike the OV-D and OV-S methods, Open-Vocabulary Visual Grounding (OV-VG) aims to detect and locate the novel-category objects referred by natural language, establishing a more extended correspondence between language and vision. Wang et al.~\cite{OV-VG} are the first to introduce the concept of OV-VG and benchmark various models on the OV-VG task.

\subsection{Vision-Language Guided Robotic Grasping}
\par With the development of the large language models~\cite{BERT,Patel-ICLR-2023} and multi-modal models~\cite{Grounding-DINO,CLIP}, vision-language guided robotic grasping has become a cutting-edge research topic. Xu et al.~\cite{VLAGrasp} access the probability of each grasping pose by jointly model action, language, and vision. Tang et al.~\cite{Tang-IROS-2023} design a task-oriented grasp method to address the challenge of task grounding in addition to object grounding. Sun et al~\cite{Sun-IROS-2023} employ CLIP~\cite{CLIP} for object localization and 6-DoF pose estimation, enhancing performance with a novel dynamic mask strategy for feature fusion. Lu et al.~\cite{VL-Grasp} propose a VG dataset called RoboRefIt to train visual-linguistic model. Despite the successful integration of visual-linguistic models into the robotic grasping task, allowing for effective manipulation of target objects within the base category, the challenge of grasping novel objects persists.
\section{Proposed Method} 
\label{sec:method}
\subsection{OVGrasping Dataset}
\par 
To empower robots with proficiency in locating and grasping both base and novel objects, we construct a new large-scale language-guided grasping dataset termed as OVGrasping, by reorganizing the RoboRefIt~\cite{VL-Grasp}, GraspNet~\cite{GraspNet} dataset, and incorporating newly collected data.
\subsubsection{\textbf{Dataset Descriptions}}
\par The OVGrapsing dataset comprises 117 categories and 63,385 instances. Instances are sourced from three distinct origins: RoboRefIt, contributing a total of 66 classes; GraspNet, enriching the dataset with 34 classes; and a simulated environment, encompassing 17 classes. Notably, we employ pybullet to craft the simulation environment, incorporating 3D models made available by OmniObject3D~\cite{omniobject3d}. Subsequently, we employ a camera positioned orthogonally to the grasping plane to capture images with dimensions of 480x640 pixels.

\par 
Specifically, The dataset is divided into two categories: the base category consists of 68 classes with 51,857 instances, and the novel category comprises 49 classes totaling 11,528 instances. The base category is used for training our framework, whereas the novel category is only utilized to evaluate the performance of open-vocabulary. Additionally, to evaluate the performance within the base category, we randomly extract 10$\%$ from the base category as the test set with 4,830 instances. Totally, the OVGrasping dataset encompasses various categories of objects and distinctly divide them into base and novel categories.

\subsubsection{\textbf{Data Annotation and Samples}}
\par 
During the annotation process, to ensure data accuracy and reliability, we hired a team of six annotation experts and three quality inspectors to check the consistency and accuracy of the annotation data. In the dataset, we provide a comprehensive description of each object, guided by the contextual semantics of the image, including color, shape, and location.
Considering that the images provided by OVGrasping potentially includes similar or identical objects, we design a new annotation format for guiding the network to better detect and locate the target object, which is inspired by RoboRefIt. Specifically, as shown in equation 1, we introduce subject-object relationship to determine the relative position, thereby distinguishing the target object from two identical objects. 

\begin{small}
\begin{equation}
\begin{split}
{Describtion} &= {<Template>} +{<Target\;Object>} \\  
&+ {<Position>} + {<Relative\;Object>}.
\label{eq3}
\end{split}
\end{equation}
\end{small}

\par
\begin{figure}
  \centering
  \includegraphics[width=1.0\linewidth]{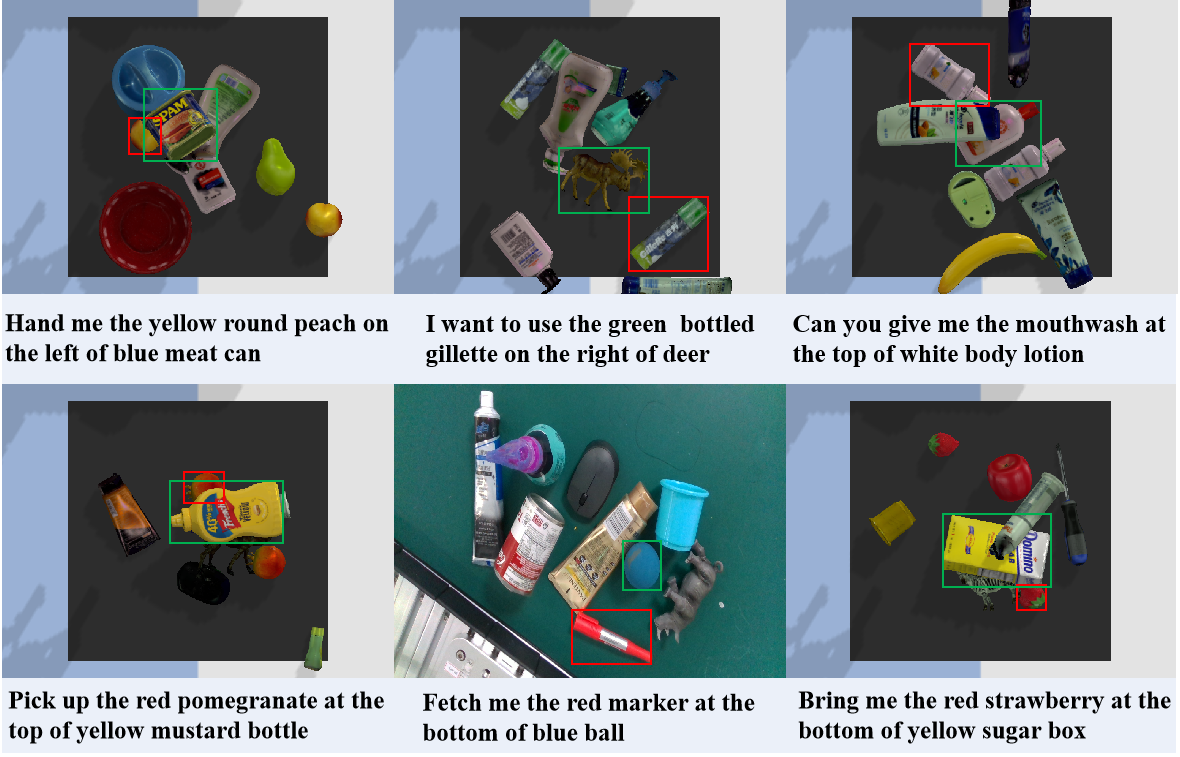}     \\
  \caption{\textbf{Samples of OVGrasping dataset.} Red boxes indicate the target objects, and green boxes denote the relative objects.}
  \label{Fig2}
\end{figure}

 During the annotation process, we randomly generate template for each target object. Annotators are tasked with providing input solely for the relative location and the corresponding object. Examples are shown in Fig.~\ref{Fig2}. We provide a detailed description to locate the target object from two identical objects. In essence, the OVGrasping dataset not only includes a plenty of base and novel classes, but also introduces the subject-object relationship to describe the relative position, providing guidance for robots to develop proficiency in grasping the target from identical objects.
\subsubsection{\textbf{Dataset Comparison}}
\par The OVGrasping dataset aims to facilitate robotic perception and grasping of base and novel objects. To fully analyze the main features of the OVGrsaping, we conduct a thorough comparison with four extensive datasets, among which the RoboRefIt is the most relevant to robotic grasping. As indicated in Table~\ref{tab1}, OVGrasping exhibits a higher average word count in comparison to RoboRefIt, accompanied by more detailed descriptions. Furthermore, compared to Sun-Spot and RoboRefIt, the OVGrasping demonstrates a notable increase in both the number of object classes and instances.
Lastly, and most importantly, the OVGrasping is an open-vocabulary (OV) dataset oriented toward robotic grasping.
\par To summarize, comparing to previous datasets, the OVGrasping has the following advantages: 1) It introduces open-vocabulary into robot grasping, clearly divides base and novel categories. 2) It uses detailed language to describe the target object, including subject-object relationship, shape, color.  3) It reorganizes data from multiple datasets, including both real and virtual. 4) It contains diverse and abundant instances, which can guide robots to perceive novel objects. 

\begin{table}[h]
\caption{Comparison of primary characteristics between previous datasets and OVGrasping dataset.}
    \centering
    \scalebox{0.9}{
        \begin{tabular}{c|c|c|c|c|c|c}
            \cmidrule(r){1-7}
            \multirow{2}{*}{Datasets} & Data & Number & Avg. & \multirow{2}{*}{OV} & Object & Data \\
            & Format & Instances & Word & & Classes & Source \\
            \cmidrule(r){1-7}
            OVGrasping & RGB & 63,385 & 10.0 & \checkmark & 117 & Real, Virtual \\
            RoboRefIt & RGBD & 50,758 & 9.5 & \(\times\) & 66 & Real \\
            \cmidrule(r){1-7}
            ScanRefer & 3D & 51,583 & 20.0 & \(\times\) & 250 & Real \\
            SUNRefer & 3D & 38,495 & 16.3 & \(\times\) & - & Real \\
            Sun-Spot & RGBD & 7,990 & 14.0 & \(\times\) & 38 & Real \\
            \cmidrule(r){1-7}
        \end{tabular}
    }
    \label{tab1}
\end{table}
\subsection{The Architecture: OVGNet}
\begin{figure*}
  \centering
  \includegraphics[width=1.0\linewidth]{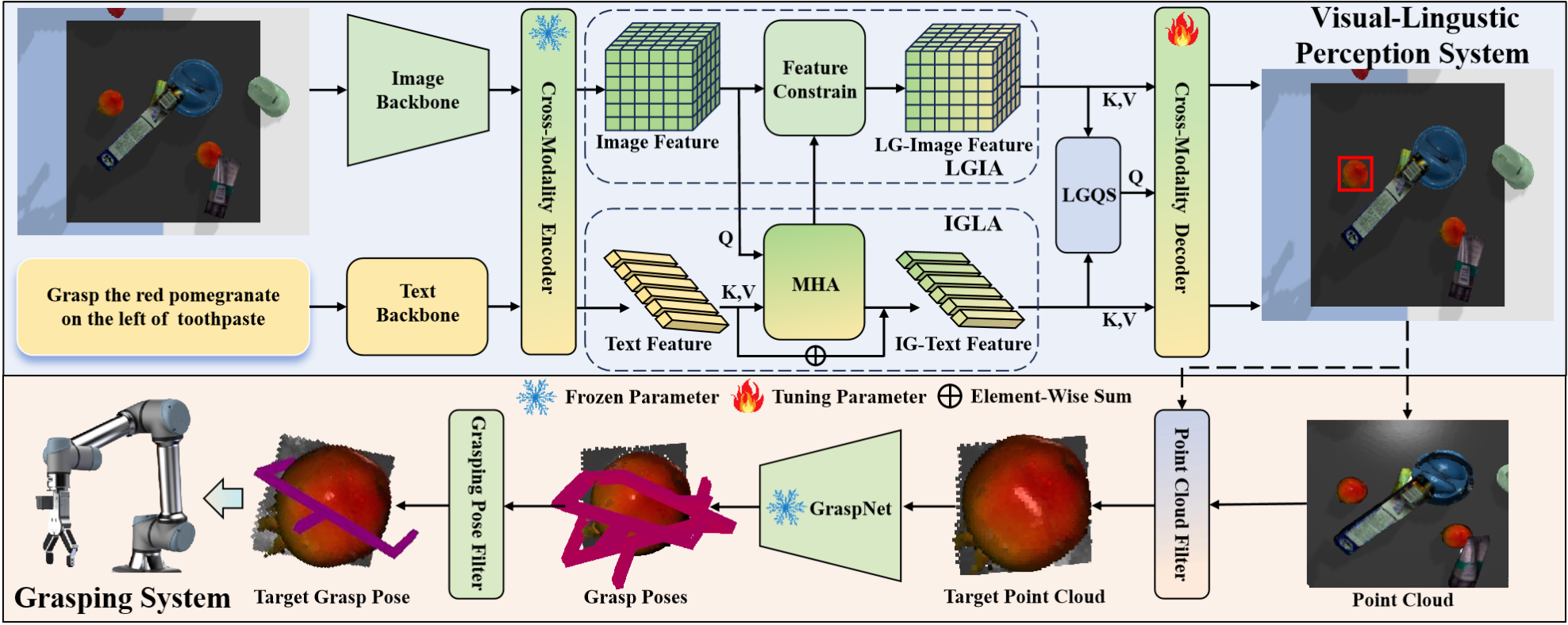}     \\
  \caption{\textbf{The overview of OVGNet.} The visual-linguistic perception system locates the target object referred by natural language. The grasping system generates grasping pose for the target object. MHA stands for multi head attention. Feature constrain represents the scaling of image feature using the constraint score. LGQS represents the language guided query selection module.}
  \label{Fig3}
\end{figure*}
\par To achieve open-vocabulary grasping, as shown in Fig.~\ref{Fig3}, we design a unified visual-linguistic grasping framework inspired by GroundingDino~\cite{Grounding-DINO} and GraspNet, which comprises two integral components: a visual-linguistic perception system and a grasping system. 
\subsubsection{\textbf{Visual-Linguistic Perception System}}
\par The system is designed to locate the base and novel target object based on language input. Given an image ${i}$ and a text description ${l}$, following the GroundingDINO, we extract image features and text features using the image backbone ${f_{i}(\cdot)}$, text backbone ${f_{l}(\cdot)}$, and cross-modality encoder ${f_{ce}(\cdot)}$, which parameters remain fixed throughout the training process.
\begin{equation}
{v_{i}}, {v_{l}} = {f_{ce}({f_{i}}(i), {f_{l}(l)})}
\label{eq2}
\end{equation}
\par Considering that the OVGrasping dataset includes complex linguistic structures such as subject-object relationship. To align the consistency between vision-language features, we design an Image Guided Language Attention (IGLA) to enhance the text feature, which is based on multi-head attention (MHA). Then, we use image feature ${v_{i}}$ as query, and text feature ${v_{l}}$ as key and value. Through multi-head attention, image features are aligned with text features, generating visual-linguistic feature ${v_{il}}$. Subsequently, we scale the vision-language features by $\alpha$, and employ a skip connection to add them to the original text feature ${v_{l}}$, yielding the image-guided text feature ${v^{'}_{l}}$. The formulation for IGLA is as follows:
\begin{equation}
{v^{'}_{l}} = \alpha{v_{il}} + {v_{l}}, \quad
{v_{il}} = softmax(\frac{{v_{i}}{v^{T}_{l}}}{\sqrt{{d_{{v_{l}}}}}}){v_{l}}
\label{eq3}
\end{equation}
Empirically, we set $\alpha$ to 0.5, where ${d_{{v_{l}}}}$ represents the dimension of ${v_{l}}$. Furthermore, to enhance visual-linguistic comprehension and heighten attention to regions described in the language within the image, we introduce a Language Guided Image Attention (LGIA) to refine the image feature ${v_{i}}$. Specifically, we employ a fully connected layer and L2 regularization to map ${v_{i}}$ and ${v^{'}_{l}}$ to the same dimensional space, obtaining ${v_{id}}$ and ${v^{'}_{ld}}$. These features are utilized to compute the constraint score ${S_{c}}$, as depicted below:

\begin{equation}
{S_{c}} = {\beta} * e^{\bigg (- \frac{{\big (1-{v_{id}(x)^{T}}{v^{'}_{ld}(x)}\big )^{2}}}{2\theta^{2}}\bigg )}
\label{eq3}
\end{equation}
where $\beta$ and $\theta$ are learnable parameters. ${v_{id}(x)^{T}}{v^{'}_{ld}(x)}$ represents the constraint score for each point ${x}$ in the vector. After obtaining the constraint score, we employ it to dampen the significance of areas in the image that are irrelevant to the language, obtaining language guided image feature ${v_{i}^{'}}$.
\begin{equation}
{v_{i}^{'}} = {\lambda}*{S_{c}}*{v_{i}} + (1-\lambda)*{v_{i}}
\label{eq4}
\end{equation}
where $\lambda$ denotes the balance parameter, which is empirically set to 0.6. Finally, following the GroundingDINO, we use language-guided query selection ${f_{ls}(\cdot)}$ to select queries ${v_{q}}$ from ${v_{i}^{'}}$. The number of queries is set to 900. Subsequently, we input ${v_{i}^{'}}$, ${v_{l}^{'}}$, and ${v_{q}}$ into cross-modality decoder ${f_{cd}(\cdot)}$, outputting the bounding box ${y}$ with the maximum score. 
\begin{equation}
{y} = argmax({f_{cd}({f_{ls}({v_{i}^{'}}, {v_{l}^{'}})}, {v_{i}^{'}, {v_{l}^{'}}})})
\label{eq4}
\end{equation}
\subsubsection{\textbf{Grasping System}}
\par The system is designed to guide the robots to grasp target object. During the experimental process, we observed that generating an accurate grasping pose for smaller objects is challenging. Therefore, we use the target bounding box obtained from visual-linguistic perception system to segment the original point cloud, obtaining the target point cloud. Subsequently, we input the target point cloud into pre-trained GraspNet to generate 6-DOF grasping poses. To enhance the success rate of grasping and filter out unsuitable postures, we use the grasp score obtained by GraspNet and grasp angle as thresholds to filter grasp poses. Following this, in the process of selecting the target grasp pose, we employ Euclidean distance to compare the distances between the center point of the bounding box and the center point of each grasp pose. The grasp pose with the shortest distance is chosen as the target grasp pose.

\section{Experiments}
\label{sec:ex}
\par In this section, we evaluate the effectiveness and rationality of our proposed framework on OVGrasping dataset and virtual environment, which includes visual-linguistic perception and grasping success rate.
\subsection{Visual-Linguistic Perception}
\subsubsection{\textbf{Training Settings}} We train the visual-linguistic system on OVGrasping dataset by using two NVIDIA Geforce RTX 3090 GPUs. The learning rate and batch size are set to 1e-5 and 16, respectively. We initialize the OVGNet parameters with GroundingDINO, which has been pre-trained on diverse datasets including COCO\cite{COCO}, O365\cite{O365}, LIVS\cite{lvis}, V3Det\cite{v3det}, Flickr30k\cite{flickr30k}, and GRIT-200K. Empirically, we set the fine-tuning epoch to 10. In the training process, we applied a random cropping for data augmentation. The longest edge of image and the max text length
are set to 1333 and 256, respectively. 
\subsubsection{\textbf{Result on OVGrasping Dataset}}
The accuracy of visual-linguistic perception is an important factor in ensuring the efficiency of open-vocabulary grasping. To fully demonstrate the effectiveness of OVGNet, we conducted a comparative experiment on the OVGrasping dataset. Following previous works~\cite{precision0.5,VL-Grasp}, we use precision@0.5 as the evaluation metric, which requires the intersection over union (IoU) between the predicted bounding box and the ground-truth box to be greater than 0.5. To ensure consistency in data usage between VL-Grasp and OVGNet, we retrained VL-Grasp on the OVGrasping dataset, in which the batch size and training epoch were set to 8 and 90, respectively.
 \par As shown in Table~\ref{tab2}, our approach exhibits a marginally lower performance compared to VL-Grasp in the base category, which may be attributed to the fewer training epochs compared with VL-Grasp. However, in contrast, our method demonstrates outstanding perceptual and generalization ability in novel category. Specifically, OVGNet achieves 47.38$\%$ improvement over VL-Grasp. Such significant improvement can be attributed to prior knowledge from the foundation model, and fine-tuning knowledge from base objects. These experimental results validate the effectiveness and 
rationality of our visual-linguistic perception system. 

\begin{table}
\caption{COMPARISON of DETECTION RESULTS with other
METHODS ON OVGRASPING.}
    \centering
    \scalebox{1.0}{
    \begin{tabular}{c|c|cc}
    \cmidrule(r){1-4}
    \multirow{3}{*}{Methods} & \multirow{3}{*}{Pre-trained Data} & \multicolumn{2}{c}{OVGrasping} \\
    \cmidrule(r){3-4}
    &  & Base & Novel \\
    \cmidrule(r){1-4}
   VL-Grasp & None & \textbf{86.08} & 18.88 \\
    \cmidrule(r){1-4}

\multirow{2}{*}{OVGNet(ours)} & COCO, O365,
LIVS, V3Det, & \multirow{2}{*}{84.22} & \multirow{2}{*}{\textbf{66.26}}  \\
    & GRIT-200K, Flickr30k. \\
     \cmidrule(r){1-4}
    
    \end{tabular}
    }
    \label{tab2}
\end{table}

\begin{table}
\caption{ Ablation study of our proposed modules on OVGrasping dataset.}
    \centering
    \scalebox{1.0}{
    \begin{tabular}{c|c|cc}
    \cmidrule(r){1-4}
   \multirow{3}{*}{IGLA} & \multirow{3}{*}{LGIA} & \multicolumn{2}{c}{OVGrasping}\\
\cmidrule(r){3-4}
   &  & Base & Novel  \\
   \cmidrule(r){1-4}
  $ \times $ & $ \times $ & 81.90 & 64.09  \\
    $ \checkmark $ & $ \times $ & 82.98 & 65.08 \\
    $ \times $ & $ \checkmark $ & 83.03 & 65.21  \\
    $ \checkmark $ & $ \checkmark $ & \textbf{84.22} & \textbf{66.26} \\
    \cmidrule(r){1-4}
    \end{tabular}
    }
    \label{tab3}
\end{table}

\subsubsection{\textbf{Visualization Experiments}}
\par
\begin{figure}
  \centering
  \includegraphics[width=1.0\linewidth]{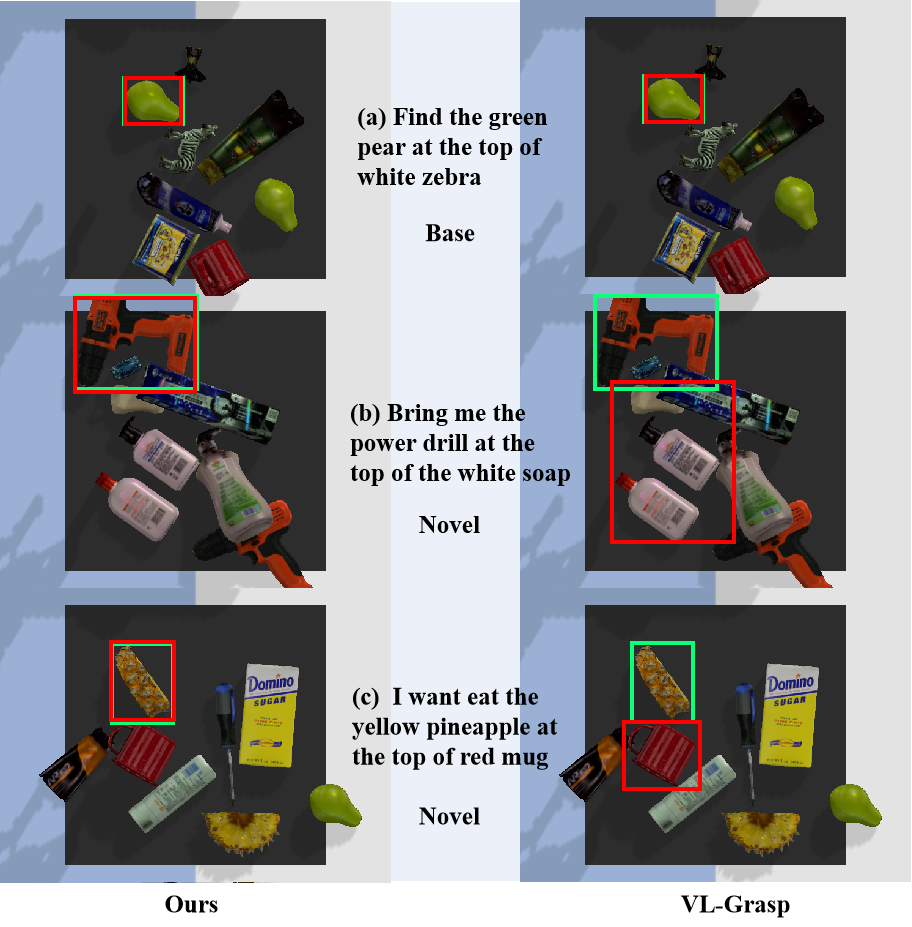}     \\
  \caption{\textbf{Visualization on OVGrasping dataset.} Green boxes indicate the ground-truth, and red boxes denote the detection results.}
  \label{Fig4}
\end{figure}

 In this section, to further analyze the characteristics of OVGNet and VL-Grasp, we conducted the visualization analysis on OVGrasping dataset. As shown in Fig.~\ref{Fig4} (a), in the base category, both VL-Grasp and OVGNet can accurately detect the target object referred by natural language. As indicated in Fig.~\ref{Fig4} (b), VL-Grasp can not efficiently detect novel objects, while OVGnet can accurately locate the novel objects. As shown in Fig.~\ref{Fig4} (c), for the target object described with subject-object relationships, VL-Grasp tends to detect the relative object that belong to the base class. In contrast, the OVGNet can effectively recognize and locate the target object. 
\subsubsection{\textbf{Ablation Study}}
\par In this section, to fully demonstrate the effectiveness of the proposed modules, we conducted the ablation studies on the OVGrasping dataset. As shown in Table~\ref{tab3}, compared to the baseline, the performance in the base category exhibits enhancements of 1.08\% and 1.13\% with the inclusion of IGLA and LGIA, respectively. In novel category, the two modules achieve an improvement with 0.99$\%$ and 1.12$\%$, respectively. With the addition of the two modules, our method outperforms the baseline by 2.32$\%$ and 2.17$\%$ in base and novel categories, respectively. The above experiments effectively validate that the integration of two modules effectively enhances the alignment between vision and language, thus improving the performance of detection accuracy in both base and novel.

\subsection{Grasping in Virtual Environment}
\subsubsection{\textbf{Grasping Setting}}
\par We constructed a grasping simulation environment based on pybullet, employing the UR5 arm and ROBOTIQ-85 to grasp target object, and using Intel RealSense L515 to capture image for the visual-language perception system. We randomly generated 135 grasping scenes to test grasping accuracy, of which 65 are used to test base and novel categories, and 70 are used to evaluate different tasks. All grasping scenes are stored in URDF format, which can be loaded by pybullet. Each scene includes 8 objects and may contain multiple identical or similar objects. To demonstrate our framework can locate the target object from identical objects, we further classify grasping scenes into single-grasping and multiple-grasping scenarios.

\subsubsection{\textbf{Grasping Test}}
\par In practical scenarios, the high trial-and-error cost for the robot hinders repeated attempts to grasp the target object. To comprehensively assess the grasping efficiency of our proposed framework, we constrain the attempts for each scene, setting the maximum attempts to 3. Specifically, once the robot successfully grasps an object in a scene, the scene automatically terminates, irrespective of whether the grasped object is the intended target or not.
\par Following the above criteria, we conducted grasping tests on 65 scenes including base and novel. As shown in Table~\ref{tab4}, our grasping framework demonstrates a remarkable success rate, requiring fewer grasping attempts. Specifically, with just one grasping attempt, our framework attains an average success rate of 44.9\% for the base objects and 34.4\% for novel objects. The success rate exhibits a consistent upward trend with additional grasping attempts. Upon reaching three attempts, our framework achieves its peak success rate of 71.2\% for base objects and 64.4\% for novel objects. In the base category, we find that our framework tends to the single grasping scenes, achieving a maximum success rate of 73.1$\%$. While in the novel category, it prefers to multiple-grasping scenes, and achieves a success rate of 65.1$\%$. The above results effectively validate the grasping performance of our framework on base and novel categories. 
\par Additionally, we select 4 simple objects and 3 hard objects to conduct the grasping test of different tasks. Each object includes 10 grasping scenes, which may contain identical objects, with the maximum of attempts as 3. As shown in Table~\ref{tab5}, in the simple task, our framework achieves an average success rate of 87.5$\%$, surpassing the average level in base category. Meanwhile, in the hard task, our framework attains an average success rate of 56.6$\%$, approaching the average level in novel class.
The above results validate the generalization of our framework in various tasks.
\begin{table}
\caption{ grasping success rate test on base and novel scenes.}
    \centering
    \scalebox{1.0}{
    \begin{tabular}{c|c|c|c|c|c|c}
    \cmidrule(r){1-7}
    \multirow{2}{*}{Attempts} & \multicolumn{3}{c}{Base} & \multicolumn{3}{c}{Novel} \\
    \cmidrule(r){2-7}
    & Single & Multi & Total & Single & Multi & Total \\
    \cmidrule(r){1-7}
    1 & 50.3 & 39.5 & 44.9 & 36.3 & 32.4 & 34.4\\
    2 & 62.1 & 53.2 & 57.7 & 54.6 & 44.5 & 49.6\\
    3 & 73.1 & 69.2 & \textbf{71.2} & 63.6 & 65.1 & \textbf{64.4}\\
    \cmidrule(r){1-7}    
    \end{tabular}
    }
    \label{tab4}
\end{table}

\begin{table}
\caption{ grasping success rate test on different tasks.}
    \centering
    \scalebox{1.0}{
    \begin{tabular}{c|c|c|c|c}
    \cmidrule(r){1-5}
    Task & Object & Base & Novel & Success Rate \\
    \cmidrule(r){1-5}
    \multirow{5}{*}{Simple} & Apple & $ \checkmark $ & & 10/10 \\
    & Orange & $ \checkmark $ & & 8/10 \\
    & Toothpaste Box & $ \checkmark $ & & 8/10 \\
    & Pear & $ \checkmark $ & & 9/10 \\
    \cmidrule(r){2-5}
     & Total & - & - & 87.5 \\
    \cmidrule(r){1-5}
    \multirow{4}{*}{Hard} & Dragon &  & $ \checkmark $& 5/10 \\
    & Power Drill & &$ \checkmark $ & 6/10 \\
    & battery & &$ \checkmark $ & 6/10 \\
    \cmidrule(r){2-5}
    & Total & - & - & 56.6 \\
    \cmidrule(r){1-5}
    
    \end{tabular}
    }
    \label{tab5}
\end{table}

\subsubsection{\textbf{Comparison with other methods}}
\par In this section, to further verify the effectiveness of our framework, we employ 8 grasping scenes provided by VLAGrasp~\cite{VLAGrasp} to test  the grasping success rate. The grasping and linguistic settings provided by VLAGrasp are different from our framework. First, in our setting, once an object is successfully grasped, regardless of whether it is the target object, the scene automatically terminates. While in VLAGrasp, the grasping process persists even if the initially grasped object is not the intended target, extending until the maximum allowed attempts are exhausted. Obviously, our grasping setting is better suited for real-world applications. Second, the language provided by VLAGrasp can refer to multiple target objects. Nevertheless, in our setting, the language must refer to single target object. Therefore, we adjust the language provided by VLAGrasp to fit our linguistic setting. The examples are shown as follows, where Ori. and Adj. represents original and adjustment language, respectively.

\begin{itemize}
    \item \textbf{Ori. :} grasp a round object.     
    \item \textbf{Adj. :} grasp a \textbf{blue} round object.
\end{itemize}

\par As shown in Table~\ref{tab6}, our framework achieves excellent success rate with fewer attempts. It is worth noting that the grasping scenes provided by VLAGrasp include 15 objects that belong to crowded scene, but our framework can still accurately locate and grasp the target objects. Such performance adequately illustrates the robust environmental adaptability of our framework.
\begin{table}
\caption{ GRASPING SUCCESS RATE test on eight scenes provided by VLAGrasp~\cite{VLAGrasp}.}
    \centering
    \scalebox{1.0}{
    \begin{tabular}{c|c|c|c|c|c}
    \cmidrule(r){1-6}
    Scenes & Target Object & Base & Novel & Success & attempts \\
    \cmidrule(r){1-6}
    1 & Racquetball &  & $ \checkmark $ &$ \checkmark $ & 1\\
    2 & Apple & $ \checkmark $ &  &$ \checkmark $ & 1\\
    3 & Blue cup & $ \checkmark $ &  &$ \checkmark $ & 1\\
    4 & Orange & $ \checkmark $ &  &$ \checkmark $ & 2\\
    5 & Banana & $ \checkmark $ &  &$ \times $ & -\\
    6 & Weiquan &  &  &$ \checkmark $ & 1\\
    7 & Theramed &  & $ \checkmark $ &$ \times $ & -\\
    8 & Pear & $ \checkmark $ &  &$ \checkmark $ & 1\\
    \cmidrule(r){1-6}
   Total & - &  - & - & 75.0 & 1.16\\
    \cmidrule(r){1-6}
    
    \end{tabular}
    }
    \label{tab6}
\end{table}
\subsubsection{\textbf{Case Analysis}}
\par
\begin{figure}
  \centering
  \includegraphics[width=1.0\linewidth]{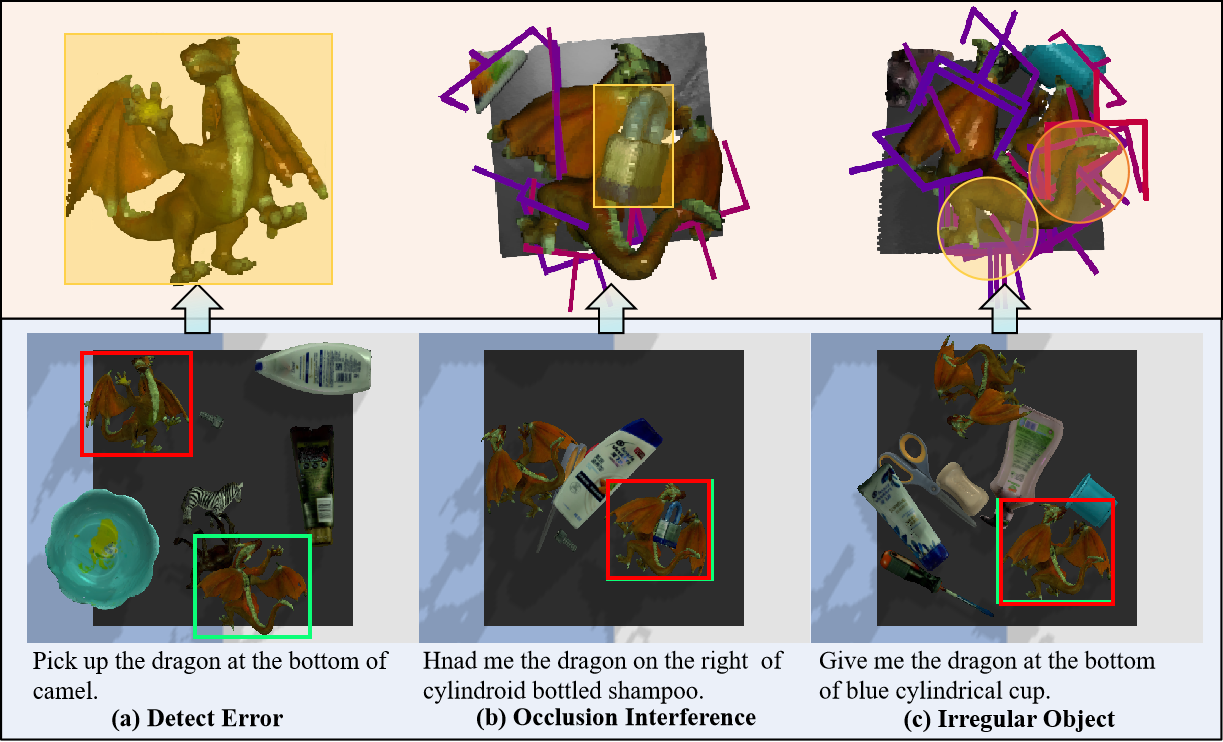}     \\
  \caption{\textbf{Case analysis.} Green boxes indicate the ground-truth, red boxes denote the predict results, and yellow area represents the defect.}
  \label{Fig5}
\end{figure}

 In this section, to better analyze the characteristics and challenges of our framework, we visualize the grasping poses of the target object and cases of grasping failures. Obviously, as shown in Fig.~\ref{Fig5} (a), when the visual-linguistic perception system outputs incorrect result, our framework is unable to locate the target object, resulting in failed grasping attempt. Furthermore, as shown in Fig.~\ref{Fig5} (b), although the target object is accurately located, there exists some occlusion in the target-level point cloud. This creates interference when generating grasping poses. Lastly, as shown in Fig.~\ref{Fig5} (c), in the attempts of grasping irregular objects with indentations or protrusions, the grasping poses are prone to these regions that are not optimal for grasping, which leads to failures in grasping.

\section{Conclusion}
\par In this paper, we present a novel approach to enhance robotic grasping by integrating open-vocabulary learning. Consequently, we contribute a challenging dataset named OVGrasping to benchmark the open-vocabulary task, and propose a unified visual-linguistic framework for open-vocabulary grasping. In the future, we will continue to explore the open-vocabulary grasping task, aiming to design an end-to-end open-vocabulary grasping system.



\bibliographystyle{IEEEbib}

\bibliography{refs}

\end{document}